\documentclass[runningheads]{llncs}

\usepackage[T1]{fontenc}
\usepackage{graphicx}
\usepackage{algorithm} 
\usepackage{algpseudocode}
\usepackage{tabularx}
\usepackage{subfigure}
\usepackage{amsmath}
\usepackage{color}
\usepackage{listings}
\usepackage{tikz}
\usetikzlibrary{arrows.meta, positioning, calc}

\begin{document}
\title{
Tailoring Optimization Approaches with Domain Knowledge in Industry Process for Meta-heuristic Development}
\titlerunning{Domain Knowledge in IPMD}

\author{Jakub Cebula\inst{1}\orcidID{0009-0002-1020-9433} \and
Konrad Gmyrek\inst{1}\orcidID{0009-0000-7206-3674} \and
Paweł B. Myszkowski\inst{1}\orcidID{0000-0003-2861-7240} \and \\
Marko Đurasević\inst{2}\orcidID{0000-0001-8732-4769} \and 
Łukasz P. Olech\inst{3}\orcidID{0000-0001-9510-626X} \and
Adam Paliszek-Saładyga\inst{1}\orcidID{0009-0002-1961-5147}}
\authorrunning{J.Cebula et al.}

\institute{
Wrocław University of Science and Technology, Faculty of Information and Communication Technology, \\ ul. Ignacego Łukasiewicza 5, 50-371 Wrocław, Poland\\
\email{pawel.myszkowski@pwr.edu.pl}
\and
University of Zagreb, Faculty of Electrical Engineering and Computing, \\ Unska ul. 3, 10000 Zagreb, Croatia
\and
Melbourne, Victoria, Australia
}

\maketitle 
\begin{abstract}
The paper presents an Industry Process for Meta-heuristic Development (IPMD) schema, a systematic methodology for applying meta- and hyper-heuristic applications to optimization problems. In this context, we present a case study of Balancing Non-dominated Tournament Genetic Programming (BNTGP) applied to solve two NP-hard multi-objective optimization problems: routing (Capacitated Vehicle Routing Problem with Time Windows) and scheduling (Multi-Skill Resource Constrained Project Scheduling Problem). The BNTGP method is based on a metaheuristic design. Its genetic programming-based representation makes BNTGP a hyper-heuristic, as it enables the exploration of a search space of heuristics for solving problems in general. The resulting heuristic can then be applied across a set of problem instances.
\end{abstract}

\section{Introduction}

The process of applying meta- and hyper-heuristics optimization to different complex problems requires multiple steps. First, researchers need to deeply understand the problem structure, constraints, choose benchmark instances, and propose a problem representation. Then they need to select, adjust, and tune the optimization method, and analyze its performance relative to other methods. 
According to the No Free Lunch (NFL) Theorem, the performance of all metaheuristic search algorithms is averaged across all possible optimization problems. However, incorporating domain knowledge can make metaheuristic design more effective in problem-aware optimization processes. Such a strategy requires experts to analyze the given problem, its instances, and the solution landscape, to define problem-specific heuristics, i.e., a solution representation, and crossover/mutation/selection operators. A hyper-heuristic can address some of these challenges by automatically adapting to different problems and scenarios, which is particularly useful for novel problems and practical applications.
This paper presents the methodology for applying the novel multi-objective Balancing Non-dominated Tournament Genetic Programming (BNTGP)~\cite{paliszek2026bntgp} to solve distinct optimization problems. The proposed Industry Process for Meta-heuristic Development (IPMD)~\cite{olech2026och} is applied to the development of the BNTGP hyper-heuristic. The IPMD methodology is a framework for the development of metaheuristics for real-world applications, and its use is demonstrated in the application of BNTGP to solving two distinct case studies: the Capacitated Vehicle Routing Problem with Time Windows (CVRPTW) and the Multi-Skill Resource Constrained Project Scheduling Problem (MS-RCPSP). This study discusses how domain knowledge is incorporated into the method and identifies common design principles that facilitate its application to new domains.

\section{Industry Process for Meta-heuristic Development}
\label{sec:ipmd} 

\begin{figure*}[t]
\centering
\scalebox{0.61}{
\begin{tikzpicture}[
    stage/.style={rectangle, draw=black, fill=blue!10, minimum width=1.7cm, minimum height=1.5cm, align=center, font=\small},
    arrow/.style={<->, thick},
    dashedarrow/.style={->, dashed},
    artifact/.style={font=\small\itshape, align=center}
]

\node[stage] (s0) {0. Problem \\ ingestion, \\ understanding};
\node[stage, right=0.6cm of s0] (s1) {1. Benchmark \\ construction \\ and selection};
\node[stage, right=0.6cm of s1] (s2) {2. Meta- or hyper- \\ heuristic design \&\\ implementation};
\node[stage, right=0.6cm of s2] (s3) {3. (Hyper) \\ Parameter tuning \\ \& benchmarking};
\node[stage, right=0.6cm of s3] (s4) {4. Performance \\ analysis  \&\\ comparisions};
\node[stage, right=0.6cm of s4] (s5) {5. Deployment \&\\ monitoring};

\foreach \i/\j in {s0/s1, s1/s2, s2/s3, s3/s4, s4/s5}
    \draw[arrow] (\i.east) -- (\j.west);

\node[artifact] at (s0.south) [yshift=-0.55cm] {$\diamondsuit$problem formula. \\ $\diamondsuit$ scope definition};
\node[artifact] at (s1.south) [yshift=-0.55cm] {$\diamondsuit$instances \\$\diamondsuit$instance analysis};
\node[artifact] at (s2.south) [yshift=-0.55cm] {$\diamondsuit$search framework \\$\diamondsuit$impl. method};
\node[artifact] at (s3.south) [yshift=-0.55cm] {$\diamondsuit$configuration(s)\\$\diamondsuit$bench. and results};
\node[artifact] at (s4.south) [yshift=-0.55cm] {$\diamondsuit$stat. test results\\ $\diamondsuit$perf. comparisons};
\node[artifact] at (s5.south) [yshift=-0.55cm] {$\diamondsuit$deployed method\\$\diamondsuit$ monitoring signals};

\draw[dashed, gray] 
    ($(s0.south)+(-2.8,-0.1)$) -- ($(s5.south)+(1.5,-0.1)$);

\node[font=\small] at 
    ($ (s0.south)!0.5!(s5.south) + (-10.0,0.1) $) {stages};
    
\node[font=\small\itshape] at 
    ($ (s0.south)!0.5!(s5.south) + (-10.0,-0.3) $) {artifacts};

\end{tikzpicture}
}
\caption{The general flow of Industry Process for Metaheuristic Development (IPMD).}
\label{fig:schema}
\end{figure*}
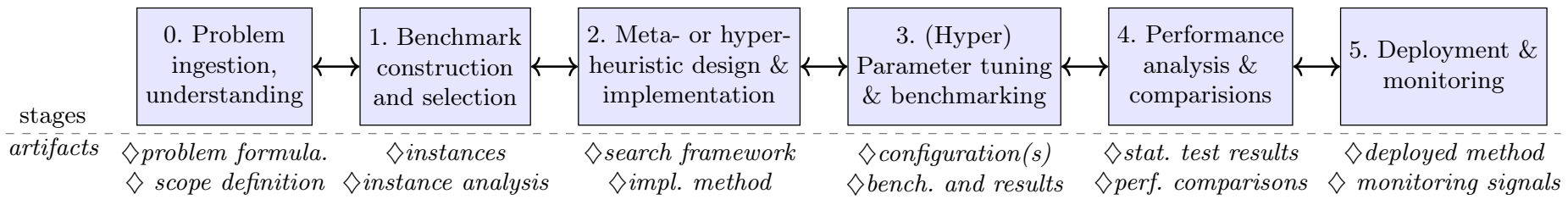

The Industry Process for Metaheuristic Development (IPMD)~\cite{olech2026och} outlines the steps required to apply meta- and hyper-heuristics to solve optimization problems. The first step of the IPMD workflow (see~Fig.\ref{fig:schema}) is \textbf{problem ingestion}, where the structure, representation, constraints and objectivities of the problem are defined. The next step involves constructing or selecting a representative \textbf{benchmark set} of the problem. Subsequently, the meta- or hyper-heuristic \textbf{method is designed and implemented} or adapted to the problem by selecting appropriate operators and defining an evaluation function. Hyperparameter \textbf{tuning} is a crucial step that directly influences the quality of the obtained results. The final stages include performance analysis and deployment of the method.

The open-source iMOPSE~\footnote{\url{https://github.com/imopse/}}~\cite{gmyrek2026imopse} library supports the IPMD process by providing researchers with problem definitions (e.g., CVRPTW, MS-RCPSP, TTP), preconfigured methods and operators, as well as tools for result analysis. iMOPSE also provides \textbf{benchmarking problem instances} for several NP-hard problems, including the MS-RCPSP standard \textit{d36} and the recently developed representative \textit{d45}~\cite{olech2026och}, which covers a wider range of benchmarking instance types, providing a more comprehensive set for developing metaheuristics for MS-RCPSP. 

\section{BNTGP for MS-RCPSP and CVRPTW: A Case Study}

The \textbf{BNTGP} is a Hyper-Heuristic for multi-objective (MO) optimization. The method combines the state-of-the-art metaheuristic Balanced Non-Dominated Tournament Genetic Algorithm (BNTGA)~\cite{Antkiewicz2024} and Genetic Programming (GP). The construction of an individual is performed by a heuristic decoder that iteratively selects the best feasible candidate action at each construction step. To determine which action is best, the method extracts features for each candidate action and combines them into a single score using an expression. Each expression is represented in the form of a tree, where the input feature values are leaves, nodes represent mathematical operations, and the root of the tree outputs a score. During optimization, a population of trees is evolved using genetic programming, where each tree is used to construct a single solution. The usage of GP allows BNTGP to automatically adjust to different problem instances. Additionally, it supports feature selection, as the features that do not contribute to the quality of results are not preserved during the evolution of trees.

The BNTGP was applied to two MO optimization problems: CVRPTW and MS-RCPSP~\cite{paliszek2026bntgp}, summarized in Table~\ref{tab:tailoring}. For \textbf{CVRPTW}, the decoder constructs vehicle routes by repeatedly selecting the next customer to visit. To describe candidate customers, a set of classical routing descriptors has been selected, such as distance, demand, remaining capacity, and time-window information.  

\vspace{-10pt}
\begin{table}[!h]
\centering
\small
\begin{tabular}{|l|c|c|}
\hline
\textbf{Aspect} & \textbf{CVRPTW} & \textbf{MS-RCPSP} \\
\hline
Type&
Routing problem &
Scheduling problem \\
\hline
Decision variable&
Customer &
Activity--resource pair \\
\hline
Decoder &
Route builder &
Schedule builder \\
\hline
Constraints &
Capacity, time windows &
Skills, task precedence \\
\hline
Basic features &
Distance, demand &
Duration, skills, costs \\
\hline
Domain features &
Regional descriptors &
Pairwise descriptors \\
\hline
Objectives (2) &
Distance, delay &
Makespan, cost \\
\hline
\end{tabular}
\caption{Summary of differences between CVRPTW and MS-RCPSP}
\label{tab:tailoring}
\end{table}
\vspace{-25pt}

\begin{figure*}
    \centering
    \includegraphics[width=0.95\textwidth]{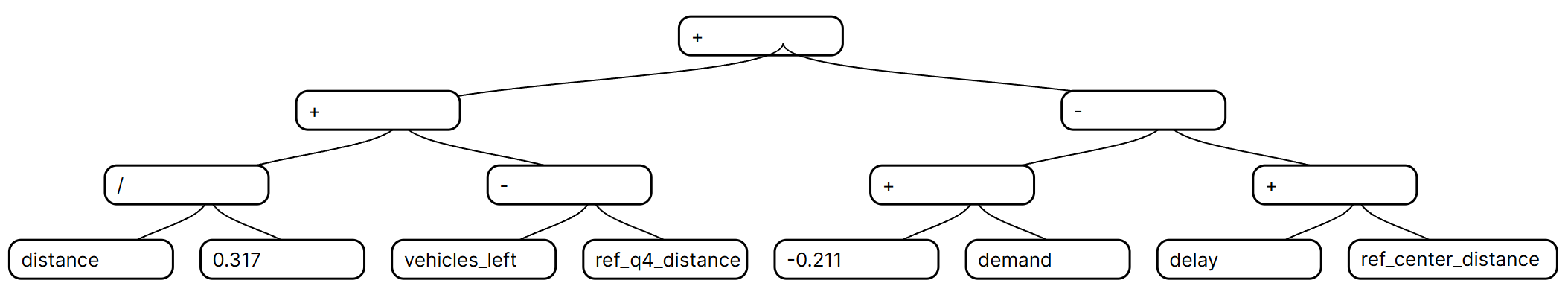}
    \caption{Example GP tree expression for CVRPTW}
    \label{fig:tree_example}
\end{figure*}

In the CVRPTW, the capacity constraint is handled directly by the decoder, such that routes are forcibly closed once vehicle capacity is reached. The number of vehicles and time window constraint violations are penalized during evaluation to discourage GP from exploiting these regions of the solution space. Figure \ref{fig:tree_example} outlines an example solution evolved by BNTGP, showcasing the use of \texttt{demand} and \texttt{vehicles\_left} features related to constraints together with \texttt{delay}. Additional region-based features that use domain knowledge provide the relative positions of customers with respect to predefined reference regions, allowing GP to exploit clustering patterns present in some benchmark instances.

For BNTGP applied to \textbf{MS-RCPSP}~\cite{paliszek2026bntgp}, candidate actions are represented as activity-resource pairs. The decoder ensures schedule feasibility with respect to precedence and skill constraints, while GP evaluates candidate pairs using features describing activity urgency, resource availability, skill compatibility, and resource costs. Additional domain-knowledge-based pair-specific descriptors were introduced to capture opportunity costs and future resource utilization.

\section{Conclusions and future works}

In this paper, the BNTGP hyper-heuristic is briefly presented in the context of IPMD as a framework for development of metaheuristics for real-world applications. Despite substantial differences between the two problems, the application of BNTGP followed a similar general pattern. The evolutionary framework remained unchanged, while domain knowledge was introduced through the design of candidate action features and decoder logic. Future work will investigate additional ways of incorporating domain knowledge into BNTGP, including automatic feature design and problem-specific feature selection. Another important direction is to study the transferability and interpretability of evolved heuristics across different instances and optimization problems.

\textbf{\\Acknowledgements:} This work has been supported by the statutory funds of the Department of Systems and Computer Networks, Faculty of Electronics, Wroclaw University of Science and Technology, and by the Croatian Science Foundation under project IP-2022-10-5398.

\bibliographystyle{splncs04}
\bibliography{bibliography}

\end{document}